%% file: st_mem.tex
\documentclass{article}

\usepackage[main,preprint]{neurips_2026}

\usepackage[utf8]{inputenc}
\usepackage[T1]{fontenc}
\usepackage{hyperref}
\usepackage{url}
\usepackage{booktabs}
\usepackage{amsfonts}
\usepackage{amsmath}
\usepackage{amssymb}
\usepackage{mathtools}
\usepackage{nicefrac}
\usepackage{microtype}
\usepackage{xcolor}
\usepackage{graphicx}
\usepackage{subcaption}
\usepackage{multirow}
\usepackage{makecell}
\usepackage{enumitem}
\usepackage{algorithm}
\usepackage{algorithmic}
\usepackage[capitalize,noabbrev]{cleveref}

\title{Linguistic Trajectory Encoding for Efficient\\Long-Horizon Spatial Memory in Embodied Agents}

\author{%
  \begin{tabular}{c}
  Xie Tianyidan\textsuperscript{1} \quad
  Shenyi Wang\textsuperscript{1} \quad
  Qiang Tang\textsuperscript{2} \quad
  Mingjie Wang\textsuperscript{3} \quad
  Zhicheng Qiu\textsuperscript{4} \\
  Xuanfu Li\textsuperscript{4} \quad
  Zhan Xu\textsuperscript{4} \quad
  Jian Yang\textsuperscript{1} \quad
  Lanjun Wang\textsuperscript{5}\textsuperscript{$\ast$} \quad
  Zili Yi\textsuperscript{1}\thanks{Corresponding authors: Zili Yi (\texttt{yi@nju.edu.cn}) and Lanjun Wang (\texttt{wanglanjun@tju.edu.cn}).} \\[4pt]
  {\normalfont \textsuperscript{1}Nanjing University \quad
  \textsuperscript{2}University of British Columbia} \\
  {\normalfont \textsuperscript{3}Zhejiang Sci-Tech University \quad
  \textsuperscript{4}Huawei Technologies Ltd. \quad
  \textsuperscript{5}Tianjin University}
  \end{tabular}
}

\begin{document}

\maketitle

\begin{abstract}
Embodied agents performing long-horizon tasks require a memory representation in which the state transitions of dynamic objects remain queryable in natural language across hours-to-days observation horizons. Existing systems either drop fine-grained motion (clip-level video-language embeddings), keep it only as raw coordinates (geometric SLAM), or organise it around immediate task context (agent working memories). None of them gives the agent a per-object timeline whose state transitions are themselves queryable in language. Our key contribution is \textbf{Linguistic Trajectory Encoding} (LTE), which compresses dynamic object motion histories via a hybrid representation combining natural language descriptions, sparse spatial anchors, and visual anchors. LTE adapts compression to motion complexity by anchoring periods without reliable observations to the last seen location, while representing motion with geometric waypoints and linguistic descriptions to preserve accuracy. To evaluate these capabilities across extended time horizons, we construct the \textbf{Spatial Memory Benchmark} (SMB) from EgoLife multi-day recordings, targeting capabilities absent in existing benchmarks: semantic trajectory retrieval and long-horizon object retrieval. On SMB, the LTE-based system achieves $45.3\%$ success in semantic trajectory retrieval and $48.7\%$ in long-horizon object retrieval, outperforming structured-memory and VLM baselines (best prior: $31.9\%$ and $34.4\%$). LTE achieves trajectory compression by factors of $8.7\times$ to $26.1\times$ with sub-second query latency on $24$\,h video. On Ego4D natural-language queries, the system reaches $28.75\%$ / $55.10\%$ R@1/R@5, $+15.80$ / $+31.30$ pts over EgoVLPv2.
\end{abstract}

%%%%%%%%%%%%%%%%%%%%%%%%%%%%%%%%%%%%%%%%%%%%%%%%%%%%%%%%%%%%%%%%%%%%%%%%%%%%%%%%
\section{Introduction}
\label{sec:intro}

\begin{figure}[t]
  \centering
  \includegraphics[width=\linewidth]{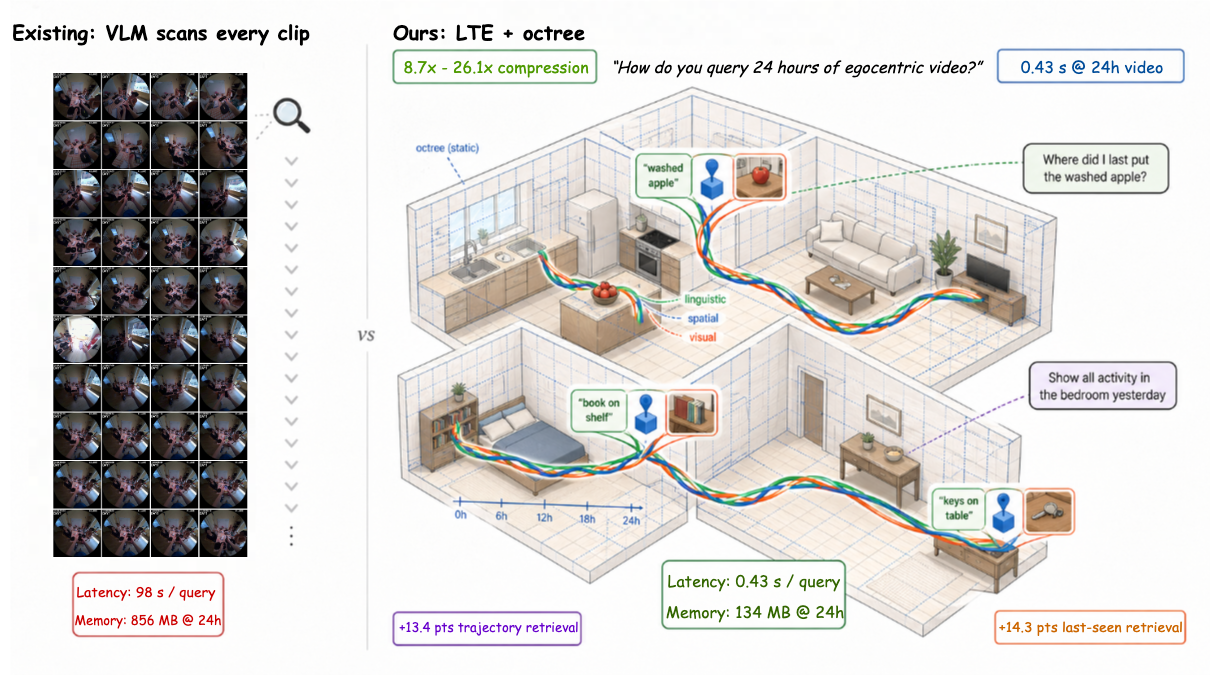}
  \caption{\textbf{Our memory system at a glance.} A 24\,h egocentric day is encoded into per-object Linguistic Trajectory Encoding (LTE) records, each braided from three channels: linguistic captions ($\mathcal{C}$, green), sparse spatial anchors ($\mathcal{A}_{\text{spatial}}$, blue), and visual anchors ($\mathcal{A}_{\text{visual}}$, terracotta). An octree provides room-level spatial pruning. Natural-language queries traverse only the relevant channel(s) and recover both the location and the moment, in $0.43$\,s, instead of $98$\,s on $24$\,h video.}
  \label{fig:teaser}
\end{figure}

Consider an embodied agent that has continuously observed a kitchen for six hours when the user asks: \emph{``Where did I last put the washed apple?''} To answer, the agent must recover, from a long observation history, an object named in language (\emph{``the apple''}), a state change described semantically (\emph{``washed''}), and a location at a specific moment in time (\emph{``where it ended up''}). A single query thus forces three retrieval modes to operate together over a memory that may span hours to days.

Existing memory systems each handle some of these modes but are limited on others. Geometric SLAM~\citep{rosinol2021kimera,khronos2024} maintains persistent 3D state over time but exposes no interface for natural-language predicates such as \emph{``washed''}. The system can tell us where an object ended up but not whether it was washed there. Video-language models~\citep{lin2022egovlp,pramanick2023egovlpv2,pei2024egovideo} match free-form text against clip-level features but lack a 3D index, \emph{``where did I last put it''} degenerates into an exhaustive temporal scan with no spatial pruning. Memory-augmented video agents~\citep{videoagent2024,amego2024} organise observations into temporal events (e.g., a single \emph{``kitchen activity''} span) but do not maintain a per-object trajectory, so the apple's individual history is folded into the broader event and lost. Task-oriented agent memories~\citep{sarch2023helper,karma2025} connect language with short-horizon spatial state but organise records around the current task. The apple, incidental to whatever task was active, leaves no queryable record at all. The shared shortfall lies in how dynamic objects are recorded: their motion is either dropped (clip embeddings), kept as raw 3D coordinates without semantic content (SLAM), or summarised only when task-relevant. \textbf{None of these systems gives the agent a per-object timeline whose state transitions are themselves queryable in language.}

\begin{figure}[t]
  \centering
  \includegraphics[width=\linewidth]{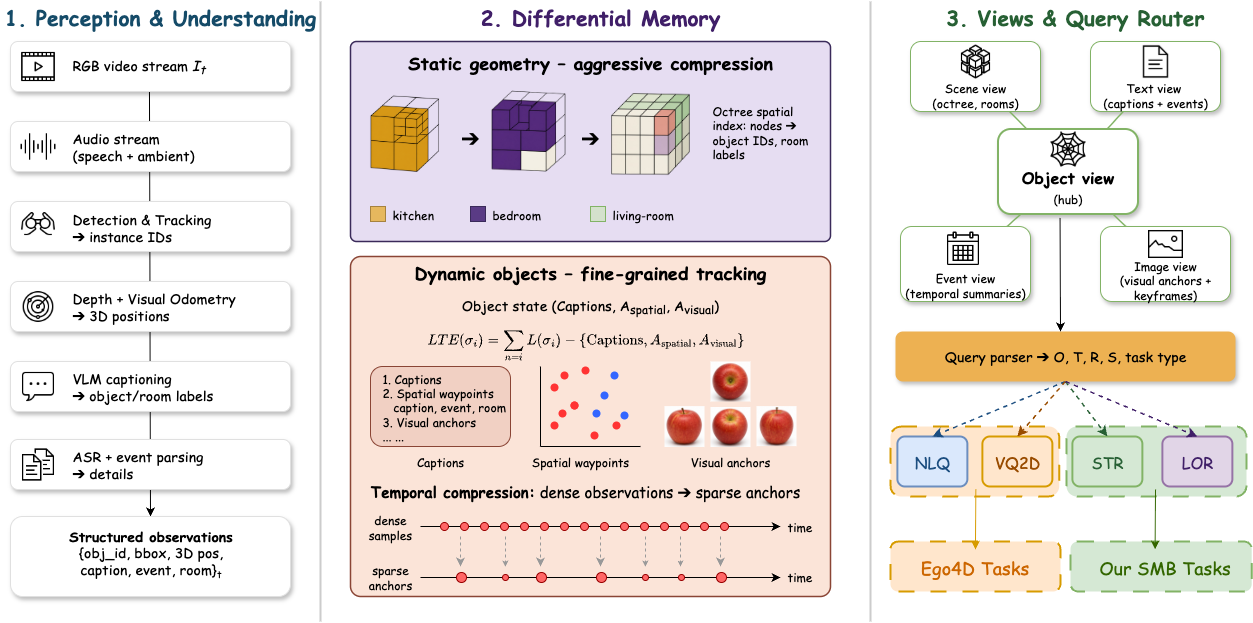}
  \caption{\textbf{LTE-centric memory architecture.} (1)~Perception (left) converts egocentric video into structured per-frame observations. (2)~The memory layer (centre) records dynamic objects with Linguistic Trajectory Encoding (captions $\mathcal{C}$, sparse spatial anchors $\mathcal{A}_{\text{spatial}}$, and visual anchors $\mathcal{A}_{\text{visual}}$) and uses an octree as a static spatial index for region pruning. (3)~Five complementary views feed a query router that dispatches to four tasks: \textbf{NLQ} and \textbf{VQ2D} from the Ego4D benchmark suite, and \textbf{STR} and \textbf{LOR} introduced in this paper as part of SMB.}
  \label{fig:framework}
\end{figure}

We address this gap with \textbf{Linguistic Trajectory Encoding} (LTE), a per-object trajectory representation designed for the dynamic-object memory problem above. For each tracked object, LTE records its history through three channels: (i)~a sequence of natural-language captions describing motion phases (e.g., \emph{``apple moved from sink to countertop''}), (ii)~sparse 3D positions retained at trajectory inflections via Douglas-Peucker simplification, and (iii)~visual crops at the same anchors for identity verification.
Each channel covers a query type that the others cannot: captions match state predicates such as \emph{``washed''}, spatial anchors answer \emph{``where''} in 3D, and visual crops verify \emph{``which apple''} across motion phases. 
Compression is adaptive: tracking-gap intervals collapse to a single anchor, simple motions retain few waypoints, and complex ones retain more, yielding compression by factors of $8.7\times$ to $26.1\times$ relative to dense per-frame storage. Answering the apple query then comes down to a single caption match against \emph{``washed''}, after which the linked 3D location and visual crop are returned (\Cref{fig:teaser}). Furthermore, to deploy LTE on hours of egocentric video, we wrap it in a memory architecture (\Cref{fig:framework}) with two supporting elements: an octree spatial index providing region pruning for queries scoped to a room or area, and five complementary views (Object, Scene, Text, Event, Image) that reuse the same per-object LTE records under different access patterns.

To evaluate the LTE-based system in the long-horizon regime that motivated it, we construct the \textbf{Spatial Memory Benchmark} (SMB) from EgoLife~\citep{yang2025egolife}: $600$ queries over multi-day recordings, with individual sessions reaching $50$\,h. SMB introduces two new tasks, Semantic Trajectory Retrieval (STR) and Long-Horizon Object Retrieval (LOR), complementing the established Ego4D Natural Language Queries (NLQ) and Visual Queries 2D (VQ2D) benchmarks~\citep{grauman2022ego4d}. On SMB, the LTE-based system achieves $45.3\%$ (STR) and $48.7\%$ (LOR), versus $31.9\%$ and $34.4\%$ for the strongest VLM baseline and $24.7\%$ and $33.8\%$ for structured-memory baselines. On Ego4D NLQ (IoU$=0.3$), it reaches $28.75\%$ / $55.10\%$ R@1/R@5, the strongest zero-shot result.

\textbf{Contributions.}
\begin{itemize}[nosep,leftmargin=1.4em]
\item \textbf{Linguistic Trajectory Encoding (LTE):} a per-object hybrid representation that records dynamic motion as language-described phases anchored to sparse 3D positions and visual crops, achieving trajectory compression by factors of $8.7\times$ to $26.1\times$ with sub-second query latency on $24$\,h video.
\item \textbf{Spatial Memory Benchmark (SMB):} $600$ queries built on EgoLife multi-day recordings, targeting two long-horizon capabilities (semantic trajectory retrieval and long-horizon last-occurrence retrieval) that are absent from existing benchmarks.
\item \textbf{Long-horizon retrieval gains:} the integrated system surpasses the strongest VLM baseline on both new SMB tasks ($+13.4$ on Semantic Trajectory Retrieval, $+14.3$ on Long-Horizon Object Retrieval), exceeds all structured-memory baselines, ranks first among zero-shot methods on the established Ego4D NLQ and VQ2D benchmarks, and approaches task-specific supervised systems on both.
\end{itemize}

%%%%%%%%%%%%%%%%%%%%%%%%%%%%%%%%%%%%%%%%%%%%%%%%%%%%%%%%%%%%%%%%%%%%%%%%%%%%%%%%
\section{Related Work}
\label{sec:related}

LTE relates to three lines of work: spatial memory representations for embodied AI, object trajectory encoding, and long-horizon video understanding. None gives a per-object timeline indexed in language. We summarise each line below and locate LTE relative to it.

\paragraph{Spatial memory and scene representation.} Static-dynamic decomposition in SLAM~\citep{schmid2022panoptic,khronos2024} separates static backgrounds from moving objects for robust localisation, and open-vocabulary 3D scene graphs~\citep{gu2024conceptgraphs,werby2024hierarchical} attach VLM-generated semantic attributes to per-object nodes. Both maintain rich spatial structure but treat each object's semantics as a static snapshot at observation time, not as a temporal sequence of state changes. Keyframe-based memories such as 3D-Mem~\citep{3dmem2025} and KARMA~\citep{karma2025} subsample observations at fixed or co-visibility-driven intervals, breaking temporal continuity. We adopt octree indexing for static geometry as a supporting element and rely on LTE for the temporal dimension that these works do not address.

\paragraph{Trajectory representation.}
Geometric trajectory compression~\citep{douglas1973algorithms} and semantic trajectory mining~\citep{yan2013semantic, zheng2015trajectory} reduce coordinate sequences to waypoints or stop-move segments, but they operate purely on numerical or categorical labels without language-grounded abstraction over motion phases. Memory-augmented video agents such as AMEGO~\citep{amego2024} and VideoAgent~\citep{videoagent2024} organize observations into temporal events or hand-object interaction tracklets, providing event-level rather than per-object linguistic structure. LTE occupies the position that none of these works fills: a per-object timeline whose successive motion phases are themselves indexed in language, anchored to sparse 3D positions and visual crops for grounding.

\paragraph{Long-horizon video.} Ego4D~\citep{grauman2022ego4d} supplies short ($\sim\!8$\,min) natural-language query tasks. EgoLife~\citep{yang2025egolife} provides continuous multi-day recordings ($300$\,h) for long-horizon evaluation. Video-language models~\citep{pramanick2023egovlpv2} and zero-shot VLMs~\citep{bai2025qwen3} lack explicit spatial structures or trajectory abstractions and lose global context across sequential clips. We address both gaps through spatial indexing plus per-object linguistic trajectories.

%%%%%%%%%%%%%%%%%%%%%%%%%%%%%%%%%%%%%%%%%%%%%%%%%%%%%%%%%%%%%%%%%%%%%%%%%%%%%%%%
\section{Method}
\label{sec:method}

\Cref{sec:problem-setup} formalizes the problem and presents the system overview. \Cref{sec:perception} describes perception components. \Cref{sec:memory-representation} details Linguistic Trajectory Encoding (the dynamic-object representation) and the octree spatial index. \Cref{sec:memory-organization} explains multi-view organization. \Cref{sec:query-processing} presents query-processing pipelines.

\subsection{Problem Setup and Architecture Overview}
\label{sec:problem-setup}

\textbf{Problem setup.}~~ An embodied agent observes an environment continuously through a video stream $V = \{I_t\}_{t=1}^T$, where $I_t \in \mathbb{R}^{H \times W \times 3}$ is the RGB frame captured at time~$t$, potentially accumulating $T > 10^5$ frames over hours to days. The agent constructs a queryable memory structure $\mathcal{M}$ supporting spatial queries specifying 3D regions, temporal queries over intervals, and semantic queries using natural language.

\textbf{System overview.}~~ The LTE-based system exploits temporal heterogeneity in spatial data. \Cref{fig:framework} illustrates the overall architecture: perception components process video streams into structured observations, memory views organize information through complementary indexing strategies, and query processing routes requests to appropriate views based on parsed constraints.

\subsection{Perception and Scene Understanding}
\label{sec:perception}

The system processes video through detection, tracking, and depth-estimation pipelines. Object detection and tracking (SAM3~\citep{sam3}) maintain instance identities across frames. Monocular depth estimation combined with visual odometry (ViPE~\citep{huang2025vipe}) reconstructs 3D positions by back-projecting 2D detections into world coordinates. A vision-language model (Qwen3-VL~\citep{bai2025qwen3}) provides room-level semantic labels (kitchen, bedroom, etc.) and object-centric motion descriptions. Speech recognition (Whisper~\citep{whisper}) transcribes audio, which a language model (Qwen3-8B~\citep{yang2025qwen3}) parses into temporal events. These perception outputs feed into memory construction.

\subsection{Linguistic Trajectory Encoding and Spatial Indexing}
\label{sec:memory-representation}

\textbf{Linguistic Trajectory Encoding for dynamic objects.}~~ Dynamic objects are the central challenge: maintaining dense, per-timestep coordinate sequences over hours of multi-object tracking is unwieldy for indexing and semantic state matching. We introduce Linguistic Trajectory Encoding (LTE) to summarise motion histories in a query-oriented hybrid form while preserving semantic accessibility and spatial grounding (\Cref{fig:framework}, lower-centre panel).

\textbf{Core insight.}~~ Many object motions admit concise semantic descriptions: \emph{``apple moved from sink to countertop''} captures essential semantics while preserving queryability. However, pure text loses the spatial precision needed for geometric queries. LTE resolves this through a hybrid representation: motion phases described linguistically, critical spatial positions retained as anchors, and visual snapshots at anchors for identity verification.

\textbf{Formal definition.}~~ For object $o_i$ with trajectory $\tau_i = \{(\mathbf{p}_t, t)\}_{t=1}^{T_i}$, LTE produces:
\begin{equation}
\text{LTE}(o_i) = (\mathcal{C}, \mathcal{A}_{\text{spatial}}, \mathcal{A}_{\text{visual}})
\end{equation}
where $\mathcal{C} = \{(c_j, [t_j^{\text{start}}, t_j^{\text{end}}])\}$ are interval captions with temporal spans, $\mathcal{A}_{\text{spatial}} = \{\mathbf{p}_{k}\}_{k \in K}$ are 3D spatial anchor positions, and $\mathcal{A}_{\text{visual}} = \{(I_{t_k}, b_{t_k})\}_{k \in K}$ are visual anchors storing frame crops and bounding boxes.

\textbf{Construction process.}~~ For each detected object $o_i$, we maintain a numeric trajectory $\tau_i$ by back-projecting bounding-box centres to 3D world coordinates. Motion state is determined by tracking continuity. Let $m_t \in \{0,1\}$ indicate whether $o_i$ is detected at timestamp $t$, and let $t^{-}(t) = \max\{t' \le t \mid m_{t'} = 1\}$. We define \textbf{tracking-gap intervals} as maximal spans where the track is missing for at least $\theta_{\text{static}}$ seconds, i.e., $m_t = 0$ and $t - t^{-}(t) \ge \theta_{\text{static}}$; during such spans we anchor the object to the last observed position $\mathbf{p}_{t^{-}(t)}$. All remaining spans are treated as motion intervals. For each interval $[t_j^{\text{start}}, t_j^{\text{end}}]$, the VLM generates caption $c_j$ describing the object's motion within scene context through visual tagging~\citep{yang2023set}.

Spatial anchors $\mathcal{A}_{\text{spatial}}$ are placed adaptively based on 3D trajectory geometry. For tracking-gap intervals, we store a single 3D position. For motion intervals, we apply Douglas-Peucker simplification~\citep{douglas1973algorithms}, retaining waypoints where the 3D trajectory deviates significantly from linear interpolation. Visual anchors $\mathcal{A}_{\text{visual}}$ store frame crops at informative timestamps: for tracking-gap intervals, we place anchors only at observable boundaries (the last detected frame before the gap and the first re-detected frame after the gap if available); for motion intervals we sample more densely to capture trajectory dynamics. This provides visual evidence for identity verification and enables visual-query matching through embedding similarity.

This hybrid representation supports multiple query types: semantic queries match caption text; spatial queries test anchor positions against query regions; visual queries compute embedding similarity against visual anchors; and temporal queries interpolate between adjacent anchors, with linguistic context providing motion semantics.

\textbf{Octree spatial indexing for static geometry.}~~ As a supporting element to LTE, we treat the octree as a static spatial partition over the reconstructed scene; nodes index object identifiers (and their anchors) to enable region-based pruning. The observed spatial extent $\Omega \subset \mathbb{R}^3$ is recursively subdivided to a maximum depth $d_{\max}$. Each node $n$ at depth $d$ covers region $\Omega_n$ and stores identifiers of objects whose 3D centres fall within its bounds. Room-level semantic labels are associated with octree regions through VLM inference on representative frames. Top-down traversal answers a query region $Q$ by pruning when $Q \cap \Omega_n = \emptyset$, collecting all descendants when $Q \supseteq \Omega_n$, and recursing otherwise.

\subsection{Memory Organization}
\label{sec:memory-organization}

We organize information into five complementary views with bidirectional links. The \textbf{Scene view} provides octree spatial indexing with VLM-inferred room-level labels at each node. The \textbf{Object view} maintains entity-centric records (category, identity, status) and links to LTE representations. The \textbf{Text view} aggregates LTE motion captions, room labels, and parsed speech transcripts behind a vector index. The \textbf{Event view} indexes temporal intervals with activity summaries derived from object-centric action captions and speech events. The \textbf{Image view} stores object-centric visual anchors $\mathcal{A}_{\text{visual}}$ plus scene-level adaptive keyframes sampled by optical-flow magnitude. Object records link to containing octree nodes, captions, and events; octree nodes link back to contained objects; events link to participants and text snippets. This supports cascaded filtering across views. Incremental updates reuse the reconstructed 3D coordinate frame, bypassing depth re-estimation (Appendix~\ref{app:memory-updates}).

\subsection{Query Processing}
\label{sec:query-processing}

The system supports four query tasks from two provenances. \textbf{Natural Language Queries (NLQ)} and \textbf{Visual Queries 2D (VQ2D)} are established Ego4D benchmarks: NLQ requires localising a temporal interval given a free-form question, VQ2D requires retrieving the most recent frame containing an object specified by a visual crop. \textbf{Semantic Trajectory Retrieval (STR)} and \textbf{Long-Horizon Object Retrieval (LOR)} are introduced as part of SMB to evaluate state-conditioned retrieval and last-occurrence retrieval, respectively, over multi-day horizons absent from Ego4D. A natural-language query $q$ is parsed by Qwen3-8B into structured constraints: objects $\mathcal{O}$, time $\mathcal{T}$, spatial regions $\mathcal{R}$, semantics $\mathcal{S}$, plus a query-type label that routes to specialised pipelines (\Cref{fig:framework}, right-bottom). All four tasks share a unified cascaded-filtering pattern (parse $\to$ route $\to$ filter $\to$ aggregate); full execution details appear in Appendix~\ref{app:query-processing}.

\textbf{NLQ.} The primary path targets the Object view: given mention $X$ we retrieve the entity record and its LTE; timeline and spatial history come from $\mathcal{C}$ and $\mathcal{A}_{\text{spatial}}$. Event constraints $Y$ trigger Event-view lookup with cross-checking against Text-view captions. Intervals are filtered by spatial (octree) and temporal (LTE timestamps) constraints, and then merged.

\textbf{VQ2D.} A DINOv2 ViT-L/14~\citep{oquab2023dinov2} embedding of the query crop is matched (cosine $\ge 0.7$) against visual anchors $\mathcal{A}_{\text{visual}}$ across all objects, temporally filtered to the most recent occurrence before $T$.

\textbf{STR.} The state description $S$ is embedded with Qwen3-Embedding-8B~\citep{qwen3embedding} and matched against LTE captions $\mathcal{C}$ in the Text view (cosine $> 0.8$), after octree pruning by $\mathcal{R}$ and $\mathcal{T}$. Matched captions resolve to object ID, temporal span, and spatial evidence.

\textbf{LOR.} Object-view entities are filtered by activity in $[T-\Delta t, T]$ and (optionally) by $\mathcal{R}$ via octree traversal; the most recent LTE anchor timestamp $t^* \le T$ and its bounding box are returned.

Query complexity depends on constraint types rather than video duration, avoiding exhaustive frame scans. Each response carries an explicit confidence label (\textsc{High}/\textsc{Medium}/\textsc{Low}) distinguishing in-window observations from extrapolated last-known positions and historical-only data, surfacing off-camera uncertainty to downstream agents (Appendix~\ref{app:uncertainty}).

%%%%%%%%%%%%%%%%%%%%%
%% EXPERIMENTS
%%%%%%%%%%%%%%%%%%%%%
\section{Experiments}
\label{sec:exp}

\subsection{Experimental Setup}

\textbf{Datasets.}~~ Ego4D~\citep{grauman2022ego4d} provides established temporal-semantic reasoning tasks. The Natural Language Queries (NLQ) task contains $5{,}462$ queries over $303$ validation videos (avg.\ $8.2$\,min). We report Recall at rank $k$ at IoU thresholds $0.3$ and $0.5$. The Visual Queries 2D (VQ2D) task provides $4{,}500$ queries across $1{,}200$ validation clips, evaluated via spatiotemporal AP (stAP), temporal AP (tAP), Success rate (IoU $\ge 0.5$), and Recovery rate. To evaluate long-horizon spatiotemporal memory, we construct the \textbf{Spatial Memory Benchmark} (SMB) from EgoLife~\citep{yang2025egolife}, which provides continuous multi-day recordings ($7$ days, $300$ total hours across $6$ participants, individual sessions reaching $50$\,h) in persistent home environments.

SMB comprises two tasks with $300$ queries each. \textbf{Semantic Trajectory Retrieval (STR)} requires locating objects based on their motion state and trajectory within specified temporal windows and semantic spatial regions; we construct $300$ queries on $232$ object instances with verified ground-truth state labels and spatiotemporal evidence. \textbf{Long-Horizon Object Retrieval (LOR)} requires retrieving the last occurrence of described objects within temporal lookback windows ranging from $2$ to $24$ hours; we construct $300$ queries with verified ground-truth evidence frames. Both tasks evaluate success: whether the system retrieves frames within the object's ground-truth temporal span with bounding-box IoU $\ge 0.3$. Construction details, IoU\,$\ge 0.5$ results, and per-horizon breakdowns appear in Appendix~\ref{app:smb} and~\ref{app:analysis}.

\textbf{Baselines.}~~ For Ego4D NLQ we compare against supervised methods EgoVLPv2~\citep{pramanick2023egovlpv2}, GroundNLQ~\citep{hou2023groundnlq}, EgoVideo~\citep{pei2024egovideo}, and OSGNet~\citep{feng2025osgnet}, plus the zero-shot method VideoAgent~\citep{videoagent2024}. For VQ2D we compare supervised VQLoC~\citep{vqloc} and PRVQL~\citep{fan2025prvql}, and zero-shot RELOCATE~\citep{khosla2025relocate}. 
For SMB we compare against two families of baselines, each probing a distinct alternative to LTE. The first tests whether a strong general-purpose VLM augmented with an open-vocabulary detector is sufficient for long-horizon retrieval: \emph{Qwen3-VL-8B-Instruct} and \emph{Qwen3-VL-235B-A22B-Instruct}~\citep{bai2025qwen3}, both paired with Grounding-DINO~\citep{groundingdino} under a $2$\,s sliding-clip protocol that scans the video at query time. The second tests whether existing structured-memory paradigms suffice without per-object linguistic trajectories: \emph{Keyframe Memory} (KFMem) follows the 3D-Mem~\citep{3dmem2025} snapshot design, using DINOv2 diversity-based keyframe selection with 3D position storage and VLM description matching, and \emph{VideoAgent}~\citep{videoagent2024} uses event-segmented memory with VLM querying.
Implementation details (models, hyperparameters, compute) and baseline protocols appear in Appendix~\ref{app:implementation} and~\ref{app:baselines}.

\subsection{Results on Ego4D Benchmarks}

\begin{table}[t]
\caption{\textbf{Ego4D NLQ validation set.} \emph{Take-away:} the LTE-based system reaches $28.75\%/55.10\%$ R@1/R@5 at IoU=0.3, improving substantially over video-language baselines (EgoVLPv2: $12.95/23.80$) and approaching task-specific supervised models. Structured object timelines and event indexing close most of the gap to supervised methods on temporal localisation.}
\label{tab:ego4d-nlq}
\centering
\small
\begin{tabular}{lcccc}
\toprule
\multirow{2}{*}{Method} & \multicolumn{2}{c}{IoU=0.3} & \multicolumn{2}{c}{IoU=0.5} \\
\cmidrule(lr){2-3} \cmidrule(lr){4-5}
& R@1 & R@5 & R@1 & R@5 \\
\midrule
\multicolumn{5}{l}{\textit{Supervised}}\\
EgoVLPv2~\citep{pramanick2023egovlpv2}  & 12.95 & 23.80 & 7.91 & 16.11 \\
GroundNLQ~\citep{hou2023groundnlq} & 27.20 & 54.42 & 18.91 & 39.98 \\
EgoVideo~\citep{pei2024egovideo} & 28.65 & 53.30 & 19.73 & 40.42 \\
OSGNet~\citep{feng2025osgnet} & \textbf{32.56} & \textbf{59.82} & \textbf{22.74} & \textbf{46.35} \\
\midrule
\multicolumn{5}{l}{\textit{Zero-shot}}\\
VideoAgent~\citep{videoagent2024} & 17.39 & 33.05 & 7.47 & 15.73 \\
Ours & 28.75 & 55.10 & 20.31 & 42.63 \\
\bottomrule
\end{tabular}
\end{table}

\Cref{tab:ego4d-nlq} presents NLQ results. At IoU=0.3 the LTE-based system reaches $28.75\%/55.10\%$ R@1/R@5, improving substantially over video-language baselines (EgoVLPv2: $12.95\%/23.80\%$) and approaching task-specific supervised models (OSGNet: $32.56\%/59.82\%$). This suggests that explicitly structured object timelines and event-indexed retrieval recover much of the benefit needed for temporal localization, while the remaining gap is consistent with supervised boundary refinement in specialized systems.

\Cref{tab:ego4d-vq2d} shows VQ2D results. We obtain the strongest zero-shot performance ($0.36$ stAP and $59.5\%$ success), slightly exceeding the prior zero-shot baseline RELOCATE ($0.33$ stAP and $58.0\%$ success). This aligns with LTE storing object-centric visual anchors that enable direct identity matching followed by temporal filtering, avoiding exhaustive frame-level scans.

\begin{table}[t]
\centering
\small
\begin{minipage}{0.46\linewidth}
\caption{\textbf{Ego4D VQ2D.} \emph{Take-away:} object-centric visual anchors enable direct identity matching, yielding the strongest zero-shot performance on all four metrics.}
\label{tab:ego4d-vq2d}
\begin{tabular}{lcccc}
\toprule
Method & stAP & tAP & Succ. & Rec. \\
\midrule
\multicolumn{5}{l}{\textit{Supervised}}\\
VQLoC  & 0.22 & 0.31 & 55.9 & 47.1 \\
PRVQL  & 0.27 & 0.35 & 57.9 & 47.9 \\
\midrule
\multicolumn{5}{l}{\textit{Zero-shot}}\\
RELOCATE  & 0.33 & 0.41 & 58.0 & 50.5 \\
Ours & \textbf{0.36} & \textbf{0.43} & \textbf{59.5} & \textbf{51.2} \\
\bottomrule
\end{tabular}
\end{minipage}\hfill
\begin{minipage}{0.50\linewidth}
\caption{\textbf{Spatial Memory Benchmark.} STR/LOR success rate (\%) on $300$ queries each. Q3VL-8B/235B+GD denote Qwen3-VL augmented with Grounding-DINO; KFMem is a 3D-Mem-style structured-memory baseline. \emph{Take-away:} the LTE-based system delivers $+13.4$ STR / $+14.3$ LOR over the strongest VLM baseline.}
\label{tab:smb}
\begin{tabular}{lcc}
\toprule
Method & STR (\%) & LOR (\%) \\
\midrule
\multicolumn{3}{l}{\textit{VLM baselines (clip-scanning)}}\\
Q3VL-8B+GD & 21.5 & 25.1 \\
Q3VL-235B+GD & 31.9 & 34.4 \\
\midrule
\multicolumn{3}{l}{\textit{Structured-memory baselines}}\\
KFMem (3D-Mem-style) & 19.8 & 33.8 \\
VideoAgent~\citep{videoagent2024} & 24.7 & 30.5 \\
\midrule
Ours & \textbf{45.3} & \textbf{48.7} \\
\bottomrule
\end{tabular}
\end{minipage}
\end{table}

\subsection{Spatial Memory Benchmark}
\label{sec:smb-results}

\Cref{tab:smb} presents SMB results. On STR, the LTE-based system achieves $45.3\%$ success, outperforming VLM and structured-memory baselines (Q3VL-235B+GD: $31.9\%$, VideoAgent: $24.7\%$, KFMem: $19.8\%$). KFMem achieves reasonable LOR ($33.8\%$) through spatial keyframe lookup but fails on STR ($19.8\%$) because it cannot represent motion histories. VideoAgent's event-based memory improves over raw VLM ($+3.2$ STR / $+5.4$ LOR) but lacks trajectory-level abstraction. Our advantage on STR ($+20.6$ over KFMem) confirms that trajectory encoding is essential for semantic state queries.

On LOR, the LTE-based system achieves $48.7\%$, outperforming all baselines. Per-horizon analysis (Appendix~\ref{app:analysis}) shows that our performance degrades mildly as the lookback window grows ($51.2\%$ at $2$\,h to $46.8\%$ at $24$\,h), whereas VLM baselines drop more sharply (Q3VL-235B+GD: $39.1\%$ to $29.3\%$). At the stricter IoU\,$\ge 0.5$ threshold the gap is preserved ($+12.4$ STR, $+13.5$ LOR; full table in Appendix~\ref{app:analysis}), confirming the result is not an artefact of a lenient threshold.

\begin{figure}[!t]
  \centering
  \includegraphics[width=0.95\linewidth]{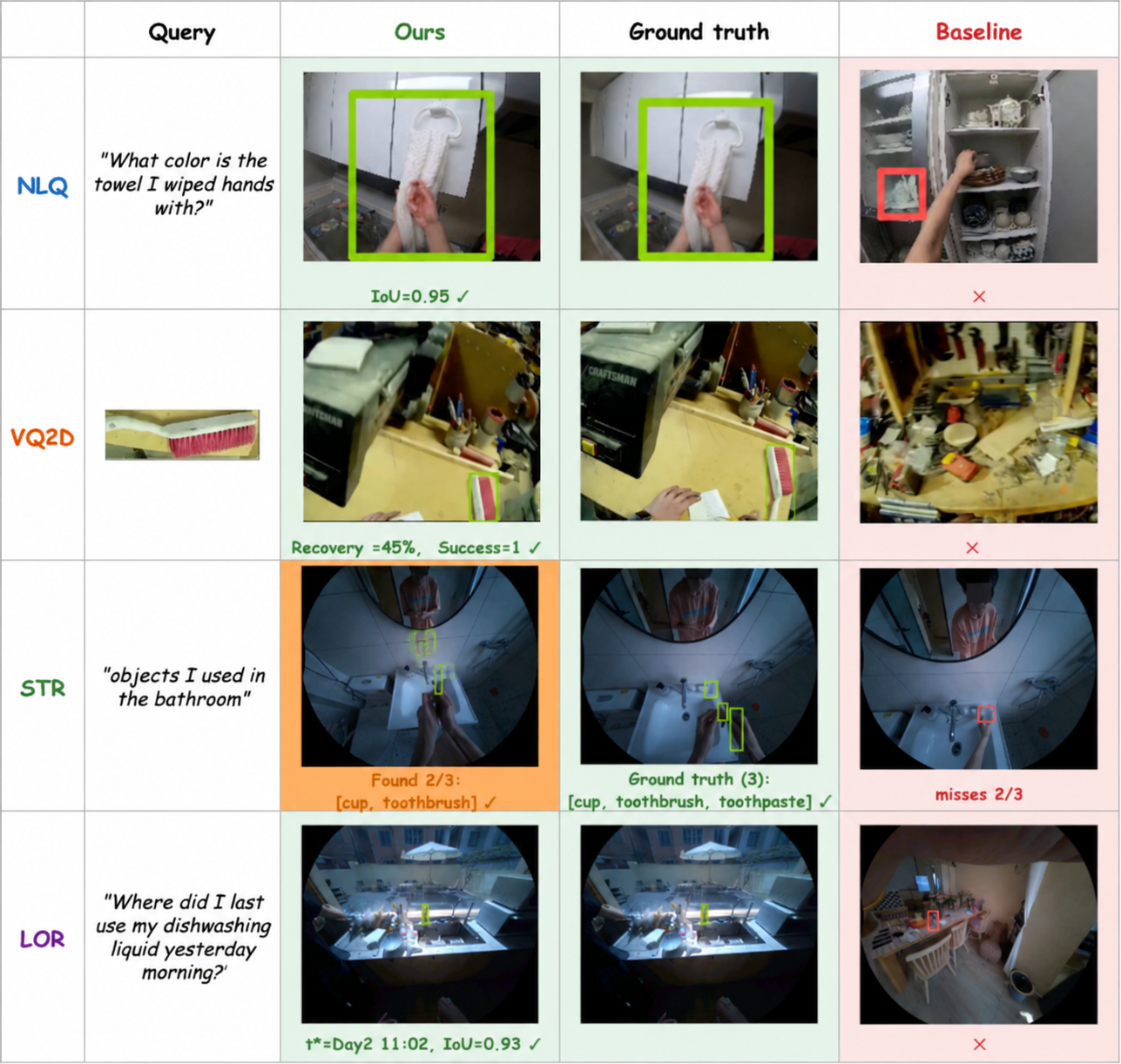}
  \caption{\textbf{Qualitative examples across the four tasks.} Each row compares our system against the strongest competing method with one success and one diagnostic failure/partial case. \textcolor[HTML]{1F7A1F}{Green}=success, \textcolor[HTML]{D69A2F}{amber}=partial, \textcolor[HTML]{B02020}{red}=failure. \emph{Take-away:} most failures originate upstream of LTE rather than from the memory representation; tracking ID-switches alone account for $34\%$ of STR errors (Appendix~\ref{app:errors}).}
  \label{fig:qualitative}
\end{figure}

\Cref{fig:qualitative} presents qualitative examples across the four tasks, including success and partial/failure cases. Common failure modes are dominated by tracking errors (ID switches and lost tracks); a full categorisation appears in Appendix~\ref{app:errors}.

\paragraph{Efficiency analysis.}
\label{sec:efficiency}
\Cref{tab:efficiency} reports memory footprint and per-query latency on a single A800~GPU. Memory grows sub-linearly ($45 \to 134$\,MB for $12\times$ more video) because LTE compression becomes more effective as tracking gaps lengthen. Query latency remains sub-second ($0.43$\,s at $24$\,h) via octree pruning, while the VLM baseline scales linearly ($98.3$\,s at $24$\,h). For $300$ SMB queries on $24$\,h video, our total query time is $2.2$\,minutes vs.\ $8.2$\,hours for the VLM. End-to-end timing breakdown is in Appendix~\ref{app:timing}.

\begin{table}[t]
\centering
\small
\caption{\textbf{Efficiency on a single A800~GPU.} Memory footprint and per-query latency vs.\ video duration. Compression ratio compares LTE-only storage against a dense per-frame trajectory store. \emph{Take-away:} memory grows sub-linearly and query latency stays sub-second up to $24$\,h, while the VLM baseline scales linearly and crosses $1.5$\,min/query.}
\label{tab:efficiency}
\setlength{\tabcolsep}{4pt}
\begin{tabular}{lccccccc}
\toprule
\multirow{2}{*}{Dur.} & \multirow{2}{*}{\#Obj.} & Dense & Total & LTE & \multirow{2}{*}{Compr.} & Ours & Q3VL-235B+GD \\
                     &                          & (MB)  & (MB)  & (MB) &                          & (s/q) & (s/q) \\
\midrule
$2$\,h  & $\sim$18  & 71   & 45  & 8.2  & $8.7\times$  & 0.15 & 8.2 \\
$6$\,h  & $\sim$35  & 214  & 73  & 16.1 & $13.3\times$ & 0.24 & 24.5 \\
$12$\,h & $\sim$52  & 428  & 98  & 23.5 & $18.2\times$ & 0.32 & 49.1 \\
$24$\,h & $\sim$78  & 856  & 134 & 32.8 & $26.1\times$ & 0.43 & 98.3 \\
\bottomrule
\end{tabular}
\end{table}

\paragraph{Ablation studies.}
\label{sec:ablations}
\Cref{tab:ablation} reports a fine-grained ablation. Removing all of LTE causes the largest STR drop ($-16.5$); removing only text captions accounts for $-11.8$ STR, isolating linguistic abstraction as the dominant contributor. Removing only visual anchors drives the largest VQ2D loss ($-6.1$); removing the octree alone reduces STR/LOR by $-3.2/-3.5$. LOR is approximately invariant to LTE (relies solely on Object view + octree), validating the modular design. The three LTE channels combine super-additively on STR: keep-only-one variants give individual gains of $+7.4$ (text), $+2.7$ (spatial), and $+1.3$ (visual) over the no-LTE baseline (Appendix~\ref{app:analysis}, Table~\ref{tab:keep-one}), summing to $+11.4$, while the full system delivers $+16.5$. The $+5.1$-pt gap reflects channel interaction: each channel becomes more useful in the presence of the others, since state-conditioned queries simultaneously require the linguistic predicate, the 3D anchor, and the visual identity check.

\begin{table}[t]
\centering
\small
\caption{\textbf{Fine-grained ablation.} Removing one component at a time across NLQ R@5 ($\%$), VQ2D Success ($\%$), SMB-STR Success ($\%$), and SMB-LOR Success ($\%$). $\Delta$ is the change relative to the full system. \emph{Take-away:} text captions drive STR ($-11.8$), visual anchors drive VQ2D ($-6.1$), the octree drives LOR ($-3.5$); LOR is invariant to LTE.}
\label{tab:ablation}
\setlength{\tabcolsep}{4.5pt}
\begin{tabular}{l@{\hskip 6pt}cccc@{\hskip 10pt}cccc}
\toprule
& \multicolumn{4}{c}{Absolute (\%)} & \multicolumn{4}{c}{$\Delta$ vs.\ Ours} \\
\cmidrule(lr){2-5}\cmidrule(lr){6-9}
Configuration & NLQ & VQ2D & STR & LOR & NLQ & VQ2D & STR & LOR \\
\midrule
Ours (full)            & 55.10 & 59.5 & 45.3 & 48.7 & --   & --   & --    & --   \\
w/o text captions      & 53.40 & 58.9 & 33.5 & 48.7 & $-1.70$ & $-0.6$ & $-11.8$ & $0.0$ \\
w/o spatial anchors    & 54.30 & 57.1 & 41.2 & 48.7 & $-0.80$ & $-2.4$ & $-4.1$  & $0.0$ \\
w/o visual anchors     & 54.70 & 53.4 & 43.9 & 48.7 & $-0.40$ & $-6.1$ & $-1.4$  & $0.0$ \\
w/o LTE (all removed)  & 52.31 & 51.8 & 28.8 & 48.7 & $-2.79$ & $-7.7$ & $-16.5$ & $0.0$  \\
w/o octree             & 54.10 & 58.7 & 42.1 & 45.2 & $-1.00$ & $-0.8$ & $-3.2$  & $-3.5$ \\
w/o LTE \& octree      & 51.80 & 51.2 & 26.3 & 44.9 & $-3.30$ & $-8.3$ & $-19.0$ & $-3.8$ \\
\bottomrule
\end{tabular}
\end{table}

%%%%%%%%%%%%%%%%%%%%%
%% DISCUSSION
%%%%%%%%%%%%%%%%%%%%%
\section{Discussion}
\label{sec:discussion}

\paragraph{Conclusion.} We introduced Linguistic Trajectory Encoding (LTE), a per-object hybrid trajectory representation that records dynamic motion as language-described phases anchored to sparse 3D positions and visual crops, together with the Spatial Memory Benchmark (SMB) that operationalises long-horizon spatial memory evaluation. The empirical core of the paper supports the framing that language at the level of motion phases, rather than at the level of clips or static object attributes, is an underused middle layer between geometric SLAM and video-language models. Text captions are the dominant single channel: the leave-one-out drop from removing them ($-11.8$) is comparable to removing all three LTE channels jointly ($-16.5$), and full LTE exceeds the sum of single-channel keep-only contributions by $+5.1$ points. Extending the same encoding strategy to articulated part dynamics, integrating with multi-camera observation fusion, and coupling LTE records with downstream planners are natural next steps.

\paragraph{Limitations.} Although LTE provides a per-object linguistic timeline for dynamic-object memory and SMB establishes a long-horizon evaluation protocol for spatial memory, several directions remain for future expansion. First, objects are tracked as atomic entities, which captures whole-object trajectories but leaves part-level dynamics such as door angles or drawer extension to future work. Augmenting LTE with an articulated-tracking layer is a natural extension, consistent with the design choice in prior object-centric memory systems~\citep{gu2024conceptgraphs,3dmem2025,karma2025}. Second, long-horizon stress tests are performed on EgoLife home environments, the only public source of multi-day continuous recordings. Extending to office, warehouse, and outdoor recordings would broaden cross-environment evaluation, with Ego4D currently supplying the cross-environment axis at shorter horizons in the present submission.

\begin{ack}
This work was supported by the National Natural Science Foundation of China (Grant No.~62406134), Jiangsu Provincial Science \& Technology Major Project (Grant No.~BG2024042), and the Suzhou Key Technologies Project (Grant No.~SYG2024136).
The authors declare no competing interests.
\end{ack}

\bibliographystyle{plainnat}
\bibliography{example_paper}

\newpage
\appendix
\input{appendix}

\end{document}

%% file: appendix.tex
% This file is included from st_mem.tex via \input{appendix}.
% It must NOT contain a preamble or \begin{document}.

%%%%%%%%%%%%%% SMB Construction %%%%%%%%%%%%%%
\section{Spatial Memory Benchmark Construction}
\label{app:smb}

\subsection{Data Source}

We construct the Spatial Memory Benchmark (SMB) from EgoLife~\citep{yang2025egolife}\footnote{\url{https://egolife-ai.github.io/}}, a comprehensive 300-hour egocentric dataset collected from six participants living together for seven days in a shared house environment. Each participant wore Meta Aria glasses recording egocentric video at $1408\times1408$ resolution. The dataset captures diverse daily activities including cooking, social interactions, housekeeping, and leisure, with rich annotations including visual--audio dense captions at various temporal granularities.

\subsection{Annotation Protocol}

Three annotators with backgrounds in computer vision and video understanding participated in the benchmark construction. Annotators were trained on the task requirements and EgoLife data structure before beginning annotation. The Semantic Trajectory Retrieval task required approximately $20$ hours of annotation per annotator ($60$ hours total), while the Long-Horizon Object Retrieval task required approximately $15$ hours per annotator ($45$ hours total), with annotators working in parallel on different video segments.

\subsection{Semantic Trajectory Retrieval Task}

\paragraph{Query construction.}
Each query specifies an object state description, optional spatial constraints, and optional temporal constraints. Annotators construct queries by combining SAM3-generated object tracking trajectories with manual verification to determine precise spatiotemporal positions of target objects. For objects where SAM3 tracking fails (e.g., due to occlusion or re-identification errors), annotators skip annotation for those instances to ensure ground-truth reliability. The skip rate was $11.3\%$ of candidate objects; $73\%$ of skipped objects appeared at $<\!40\times40$ pixels (below reliable detection for both SAM3 and Grounding-DINO, ensuring the same difficulty floor for VLM baselines).

Each query involves either one or two objects. Among the $300$ queries, $187$ ($62.3\%$) involve a single object and $113$ ($37.7\%$) involve two objects (e.g., ``put the book into the backpack''). Object states are annotated based on the Visual--Audio Dense Captions provided by EgoLife, combined with manual observation to create object-centric event annotations.

\paragraph{Spatial and temporal constraints.}
For spatial constraints, we use semantic region descriptions: Bedroom, Living room, Kitchen, Dining room, Bathroom, and Other (outdoor locations and transitional spaces). For temporal constraints, we map the original timestamps from EgoLife to semantic descriptions (e.g., ``09:00'' $\rightarrow$ ``morning''). Among the $300$ queries, $156$ ($52.0\%$) include spatial hints, $168$ ($56.0\%$) include temporal hints, and $89$ ($29.7\%$) include both.

\paragraph{Ground truth establishment.}
For each query, annotators identify object instances matching the state description and annotate temporal spans $[t_{\text{start}}, t_{\text{end}}]$. The temporal span begins when the object enters the specified state and ends when either: (i) the object transitions to a different state, or (ii) the object becomes undetectable for at least $2$ seconds. Bounding boxes are annotated at representative frames within each valid span. Inter-annotator agreement (Fleiss' $\kappa$) is $0.71$ for object identification and $0.65$ for temporal boundary annotation.

\subsection{Long-Horizon Object Retrieval Task}

\paragraph{Query construction.}
This task requires retrieving the last occurrence of described objects within specified temporal lookback windows. To ensure accurate ``last occurrence'' annotation, we extract video segments of $2$\,h, $6$\,h, $12$\,h, and $24$\,h duration sorted chronologically, and annotators observe these segments in reverse temporal order (newest to oldest).

\begin{table}[h]
\centering
\caption{Long-Horizon Object Retrieval query distribution by lookback horizon.}
\label{tab:smb-lor-distribution}
\begin{tabular}{lccc}
\toprule
Lookback Window & Count & With Spatial Hint & Avg.\ Duration \\
\midrule
$2$ hours  & $75$ & $32$ ($42.7\%$) & $3.2$--$4.8$\,s \\
$6$ hours  & $75$ & $29$ ($38.7\%$) & $3.5$--$5.0$\,s \\
$12$ hours & $75$ & $31$ ($41.3\%$) & $3.1$--$4.6$\,s \\
$24$ hours & $75$ & $28$ ($37.3\%$) & $3.8$--$5.2$\,s \\
\midrule
\textbf{Total} & $300$ & $120$ ($40.0\%$) & $3.4$--$4.9$\,s \\
\bottomrule
\end{tabular}
\end{table}

Among all queries, $40.0\%$ include spatial hints, $45.3\%$ include temporal hints, and $22.7\%$ include both.

\paragraph{Ground truth annotation.}
Annotators mark the last occurrence frame $t^*$ and corresponding bounding box $b^*$ within each lookback window. The annotated temporal span typically ranges from $3$ to $5$ seconds. Inter-annotator agreement is $0.78$ for last-occurrence timestamp (within $3$-second tolerance) and $0.72$ for bounding box (IoU $\ge 0.5$).

%%%%%%%%%%%%%% Implementation Details %%%%%%%%%%%%%%
\section{Implementation Details}
\label{app:implementation}

\subsection{Compute}
All experiments are run on an NVIDIA A800 GPU ($80$\,GB).

\subsection{Model Specifications}

We use SAM3~\citep{sam3} for instance segmentation and tracking, ViPE~\citep{huang2025vipe} for 3D reconstruction, Qwen3-8B~\citep{yang2025qwen3} for query parsing, Qwen3-VL-8B-Instruct~\citep{bai2025qwen3} for room labeling and captioning, Qwen3-Embedding-8B~\citep{qwen3embedding} for text embeddings, DINOv2~\citep{oquab2023dinov2} for visual embeddings, SigLIP2~\citep{tschannen2025siglip2} for text--image matching, and Whisper~\citep{whisper} for speech transcription. \Cref{tab:models} lists all models with their roles.

\begin{table}[h]
\centering
\caption{Model specifications.}
\label{tab:models}
\begin{tabular}{lll}
\toprule
Component & Model & Role \\
\midrule
Object detection & SAM3 (ViT-H) & Instance segmentation \\
Object tracking & SAM3 tracker & Cross-frame association \\
Point cloud & ViPE (ViT-L) & 3D reconstruction \\
VLM (scene) & Qwen3-VL-8B-Instruct & Room-level semantic labeling, captions \\
LLM (parsing) & Qwen3-8B & Query parsing, event extraction \\
Text embedding & Qwen3-Embedding-8B & Semantic similarity \\
Visual embedding & DINOv2 ViT-L/14 & Visual anchor matching \\
Text--image matching & SigLIP2-B/16 & Cross-modal retrieval \\
Speech recognition & Whisper-large-v3 & Audio transcription \\
\bottomrule
\end{tabular}
\end{table}

\subsection{Hyperparameters}

\paragraph{Octree spatial indexing.}
Following the EgoLife house layout, we configure the octree with maximum depth $d_{\max} = 8$, minimum node size $0.1\text{m} \times 0.1\text{m} \times 0.1\text{m}$, and root node covering $20\text{m} \times 20\text{m} \times 5\text{m}$.

\paragraph{Linguistic Trajectory Encoding.}
\begin{itemize}[nosep,leftmargin=*]
    \item Douglas--Peucker simplification tolerance: $0.15$\,m (perpendicular distance)
    \item Minimum anchor spacing: $0.25$\,seconds
    \item Maximum anchor spacing: $10$\,seconds
    \item Tracking-gap threshold $\theta_{\text{static}}$: $2$\,seconds
    \item Caption generation: per motion phase transition
\end{itemize}

\paragraph{VLM-based Adaptive Scene Understanding.}
To achieve scalable room-level semantic labeling for long-horizon egocentric videos, we implement an adaptive sliding-window mechanism powered by Qwen3-VL-8B. The system processes video segments in $3$-second windows (sampled at $3$\,FPS) to simultaneously infer the semantic room label $\mathcal{L} \in \{\text{Bedroom, Living room, Kitchen, Dining room, Bathroom, Other}\}$ and identify salient keyframes. The VLM is prompted to perform both tasks jointly (see Appendix~\ref{app:prompts}). To minimize redundant computation during prolonged stays in static environments, we employ an \textbf{exponential backoff stride strategy}: given $k$ consecutive windows with a consistent label, the subsequent temporal stride is $s_{k+1} = \min(2^{\lfloor k/2 \rfloor} \cdot s_0, s_{\max})$, where $s_0 = 3$\,s and $s_{\max} = 15$\,s. Keyframes are selectively committed based on three criteria: (i) room entry frames, (ii) room exit frames, (iii) salient in-room frames. Upon detecting a label shift, the system resets the stride to $s_0$ and performs a \textbf{temporal backtracking search} within $[t-s, t]$ to precisely localize the transition boundary.

\begin{algorithm}[h]
\caption{Adaptive Semantic Scene Indexing}
\label{alg:adaptive_vlm}
\begin{algorithmic}[1]
\STATE {\bfseries Input:} Video stream $\mathcal{V}$, base stride $s_0=3$\,s, max stride $s_{\max}=15$\,s.
\STATE {\bfseries Initialize:} Current time $t \leftarrow 0$, stability counter $k \leftarrow 0$, $\mathcal{L}_{prev} \leftarrow \emptyset$.
\WHILE{$t < \text{Duration}(\mathcal{V})$}
      \STATE $\mathcal{L}_t, \{f_{key}\} \leftarrow \text{VLM}(\text{Clip}(t, t+3\text{s}))$ \COMMENT{Infer room label and extract keyframes}
      \IF{$\mathcal{L}_t = \mathcal{L}_{prev}$}
         \STATE $k \leftarrow k + 1$
         \STATE $s \leftarrow \min(2^{\lfloor k/2 \rfloor} \cdot s_0, s_{\max})$ \COMMENT{Exponential backoff for stability}
      \ELSE
         \IF{$\mathcal{L}_{prev} \neq \emptyset$}
            \STATE $t_{boundary} \leftarrow \text{BacktrackSearch}(\mathcal{V}, t-s, t)$ \COMMENT{Locate precise transition}
            \STATE $\text{CommitRoomNode}(t_{boundary}, \mathcal{L}_t)$
         \ENDIF
         \STATE $s \leftarrow s_0, k \leftarrow 0, \mathcal{L}_{prev} \leftarrow \mathcal{L}_t$
      \ENDIF
      \STATE $\text{SaveKeyframes}(\{f_{key}\})$
      \STATE $t \leftarrow t + s$
\ENDWHILE
\end{algorithmic}
\end{algorithm}

\paragraph{Query processing thresholds.}
\begin{itemize}[nosep,leftmargin=*]
    \item Visual similarity threshold (VQ2D): $0.7$ (cosine similarity)
    \item Semantic similarity threshold (STR): $0.8$ (cosine similarity)
    \item Object semantic fallback threshold: $0.75$
\end{itemize}

%%%%%%%%%%%%%% End-to-End Timing %%%%%%%%%%%%%%
\section{End-to-End Timing}
\label{app:timing}

\Cref{tab:timing} reports the full pipeline timing breakdown from raw video to query response on a single A800~GPU.

\begin{table}[h]
\centering
\caption{\textbf{End-to-end timing.} Offline memory construction (one-time) and online per-query latency.}
\label{tab:timing}
\begin{tabular}{lccc}
\toprule
Component & $2$\,h video & $24$\,h video & Parallelizable? \\
\midrule
\multicolumn{4}{l}{\emph{Offline memory construction}}\\
SAM3 detection + tracking      & $2.8$\,h  & $33.6$\,h & Yes (per-segment) \\
ViPE 3D reconstruction         & $1.9$\,h  & $22.8$\,h & Yes (independent) \\
VLM scene labeling             & $0.4$\,h  & $3.2$\,h  & Yes (per-window) \\
VLM object captioning          & $0.8$\,h  & $8.5$\,h  & After tracking \\
LTE construction + indexing    & $0.05$\,h & $0.3$\,h  & After captioning \\
\midrule
Total (sequential)             & $5.95$\,h & $68.4$\,h & --- \\
Total (parallel pipeline)      & $\sim\!3.6$\,h & $\sim\!42$\,h & --- \\
\midrule
\multicolumn{4}{l}{\emph{Online query (per query)}}\\
Ours                           & $0.15$\,s & $0.43$\,s & --- \\
Q3VL-235B+GD                   & $8.2$\,s  & $98.3$\,s & --- \\
\bottomrule
\end{tabular}
\end{table}

Memory construction is a one-time offline cost. The break-even point relative to Q3VL-235B+GD is approximately $1{,}545$ queries on a $24$\,h video ($42$\,h $\times\,3{,}600$\,s/h $\div\,97.87$\,s saved per query). For persistent environments where agents operate continuously, this cost is paid once and all subsequent queries benefit from sub-second latency. Reducing construction time through adaptive frame skipping during stable scenes, incremental construction (\Cref{sec:memory-organization}), and lighter perception models for time-critical scenarios is an active area of work.

%%%%%%%%%%%%%% Analysis %%%%%%%%%%%%%%
\section{Per-Horizon Analysis and Additional Results}
\label{app:analysis}

\subsection{Long-Horizon Object Retrieval Breakdown}

\Cref{tab:lor-breakdown} presents detailed results across temporal horizons.

\begin{table}[h]
\centering
\caption{Long-Horizon Object Retrieval success rate (\%) by lookback horizon.}
\label{tab:lor-breakdown}
\begin{tabular}{lccccc}
\toprule
Method & $2$\,h & $6$\,h & $12$\,h & $24$\,h & Avg. \\
\midrule
Q3VL-8B+GD   & $32.0$ & $26.7$ & $22.7$ & $18.9$ & $25.1$ \\
Q3VL-235B+GD & $39.1$ & $36.5$ & $32.8$ & $29.3$ & $34.4$ \\
\midrule
Ours & $\mathbf{51.2}$ & $\mathbf{49.3}$ & $\mathbf{47.8}$ & $\mathbf{46.8}$ & $\mathbf{48.7}$ \\
\midrule
$\Delta$ (Ours vs.\ Q3VL-235B) & $+12.1$ & $+12.8$ & $+15.0$ & $+17.5$ & $+14.3$ \\
\bottomrule
\end{tabular}
\end{table}

\paragraph{Degradation analysis.}
Our system exhibits $4.4$-point degradation from $2$\,h to $24$\,h ($51.2\% \to 46.8\%$), while Qwen3-VL-235B degrades by $9.8$ points ($39.1\% \to 29.3\%$). This stability stems from octree spatial indexing and object-centric tracking: query complexity depends on spatial constraint selectivity rather than video duration. VLM baselines process videos as sequential clips, accumulating temporal discontinuities that cause progressive degradation over longer horizons.

\subsection{Stricter IoU Thresholds}

\Cref{tab:iou-strict} compares performance at IoU\,$\ge 0.3$ versus IoU\,$\ge 0.5$. The advantage holds at the stricter threshold ($+12.4$ STR, $+13.5$ LOR), confirming that results are not an artefact of a lenient IoU criterion.

\begin{table}[h]
\centering
\caption{SMB performance at stricter IoU thresholds (\%).}
\label{tab:iou-strict}
\begin{tabular}{lcccc}
\toprule
Method & STR@$0.3$ & STR@$0.5$ & LOR@$0.3$ & LOR@$0.5$ \\
\midrule
Q3VL-235B+GD & $31.9$ & $22.4$ & $34.4$ & $25.7$ \\
Ours         & $\mathbf{45.3}$ & $\mathbf{34.8}$ & $\mathbf{48.7}$ & $\mathbf{39.2}$ \\
$\Delta$     & $+13.4$ & $+12.4$ & $+14.3$ & $+13.5$ \\
\bottomrule
\end{tabular}
\end{table}

\subsection{Semantic Trajectory Retrieval by Query Type}

\begin{table}[h]
\centering
\caption{STR success rate (\%) by query type.}
\label{tab:str-breakdown}
\begin{tabular}{lccc}
\toprule
Query Type & Q3VL-235B+GD & Ours & $\Delta$ \\
\midrule
Single object queries        & $34.2$ & $48.1$ & $+13.9$ \\
Two-object queries           & $28.3$ & $40.7$ & $+12.4$ \\
With spatial hint only       & $33.5$ & $46.8$ & $+13.3$ \\
With temporal hint only      & $31.2$ & $44.2$ & $+13.0$ \\
With both hints              & $35.8$ & $49.5$ & $+13.7$ \\
\midrule
\textbf{Overall}             & $31.9$ & $\mathbf{45.3}$ & $+13.4$ \\
\bottomrule
\end{tabular}
\end{table}

Our system maintains consistent advantages across query types. Queries with both spatial and temporal hints achieve the highest performance ($49.5\%$), as combined constraints enable more precise filtering. Two-object queries show lower absolute performance due to the compounding effect of potential tracking errors across multiple entities.

\subsection{Keep-Only-One LTE Variants}

To further isolate per-component contributions, \Cref{tab:keep-one} reports performance when only \emph{one} LTE channel is retained. Individual components yield STR gains of $+7.4$ (text), $+2.7$ (spatial), and $+1.3$ (visual) over no-LTE; the full system achieves $+16.5$, exceeding the sum of individual gains ($+11.4$). This super-additive behavior indicates meaningful interaction between channels.

\begin{table}[h]
\centering
\caption{Keep-only-one LTE variants. Success/recall (\%) on each task.}
\label{tab:keep-one}
\begin{tabular}{lcccc}
\toprule
Configuration & STR & LOR & NLQ R@5 & VQ2D Succ. \\
\midrule
Full LTE              & $45.3$ & $48.7$ & $55.10$ & $59.5$ \\
Text captions only    & $36.2$ & $48.3$ & $53.80$ & $52.4$ \\
Spatial anchors only  & $31.5$ & $48.5$ & $52.90$ & $52.1$ \\
Visual anchors only   & $30.1$ & $48.4$ & $52.50$ & $56.8$ \\
No LTE                & $28.8$ & $48.7$ & $52.31$ & $51.8$ \\
\bottomrule
\end{tabular}
\end{table}

%%%%%%%%%%%%%% Error Analysis %%%%%%%%%%%%%%
\section{Error Analysis}
\label{app:errors}

We analyse $100$ randomly sampled failure cases from SMB ($50$ STR, $50$ LOR) to identify systematic error patterns.

\begin{table}[h]
\centering
\caption{Failure case categorization on SMB (\%).}
\label{tab:errors}
\begin{tabular}{lcc}
\toprule
Error Category & STR & LOR \\
\midrule
Tracking failure (ID switch, lost track) & $34$ & $28$ \\
Caption ambiguity (imprecise description) & $28$ & $12$ \\
Spatial localization error (point cloud drift) & $18$ & $24$ \\
Query parsing error & $12$ & $18$ \\
Object occlusion (partial/full) & $8$  & $18$ \\
\bottomrule
\end{tabular}
\end{table}

\paragraph{Tracking failures.}
The dominant error source ($34\%$ for STR, $28\%$ for LOR) includes: (i)~\emph{ID switches}: objects with similar appearance passing near each other occasionally cause SAM3 to swap identities; (ii)~\emph{Track loss}: objects leaving the field of view for more than $2$\,seconds may not be re-associated upon reappearance; (iii)~\emph{Small-object issues}: objects appearing at small scales (e.g., $\le 30 \times 30$ pixels) are prone to both missed detections and misidentification.

\paragraph{Caption ambiguity.}
LTE captions occasionally lack specificity for precise state matching ($28\%$ of STR failures). Generated descriptions like ``object was placed on surface'' may omit critical source-location information needed to match queries specifying motion paths.

\paragraph{Spatial localization errors.}
Running ViPE for extended durations (multiple hours) accumulates point cloud drift ($18\%$ STR, $24\%$ LOR). Monocular depth estimation and visual odometry components exhibit systematic drift over time, particularly during camera motion through doorways or between floors. Two design choices mitigate this: LTE's Douglas--Peucker simplification ($0.15$\,m tolerance) absorbs frame-level depth noise, and the octree's VLM-based room labels provide drift-robust spatial fallbacks (room labels remain correct even when metric positions drift).

\paragraph{Semantic correction (future direction).}
Tracking errors are not currently corrected automatically. The multi-view architecture supports semantic consistency checks: an LTE caption ``apple moved to countertop'' combined with a spatial anchor pointing to the bedroom flags a likely ID switch. Event view temporal context can help resolve ambiguous re-identifications. Implementing this as a post-processing pass over already-built LTE records does not require changes to the core architecture.

%%%%%%%%%%%%%% Uncertainty Quantification %%%%%%%%%%%%%%
\section{Uncertainty Quantification in Query Responses}
\label{app:uncertainty}

A single ego-camera cannot directly observe state changes that happen off-frame; if an object is moved while out of view, the system can only report the last known state. We attach explicit confidence labels to query responses to surface this uncertainty:

\begin{itemize}[nosep,leftmargin=*]
\item \textbf{\textsc{High}}: object observed within the query's temporal window; bounding box directly available.
\item \textbf{\textsc{Medium}}: object last observed before the query window; position extrapolated from the most recent LTE spatial anchor. Response includes ``last seen at [location] at [time], current location uncertain.''
\item \textbf{\textsc{Low}}: object not observed in recent history; only historical LTE data available.
\end{itemize}

This labelling is lightweight (compares observation timestamps against query timestamps) and provides users with actionable uncertainty information. Integration with external sensors (motion detectors, smart-home devices) can provide off-camera updates through the Event view, but is outside our current single-camera scope.

%%%%%%%%%%%%%% Additional Method Details %%%%%%%%%%%%%%
\section{Additional Method Details}

\subsection{Memory Update Strategy}
\label{app:memory-updates}

The memory construction pipeline processes video through parallelizable modules. Point cloud reconstruction (ViPE), object detection and tracking (SAM3), and scene-level understanding (VLM-based room-level semantic labeling) execute independently on the input video stream. Object-level description generation via VLM depends on SAM3 segmentation outputs and executes after tracking completes for each segment.

The system supports incremental addition of newly observed objects to existing memory by reusing the reconstructed 3D coordinate frame, avoiding rerunning 3D reconstruction over the full video history. For new object $o_{\text{new}}$ detected in additional frames, we re-execute detection, tracking, semantic understanding, and inter-object association to ensure consistency. However, 3D localization leverages the pre-computed point cloud through direct depth-map lookup, bypassing re-execution of depth estimation. Once $o_{\text{new}}$'s LTE representation is constructed, all memory views are updated.

\paragraph{Text materialization.}
The Text view aggregates linguistic content from three distinct sources, each serving different query patterns:

\textbf{(i) LTE motion captions:} For each object $o_i$ with trajectory encoded via LTE, motion phase intervals $\{[t_j^{\text{start}}, t_j^{\text{end}}]\}$ are associated with natural-language descriptions $\{c_j\}$ generated by the VLM. Each caption describes object behavior within spatial context (e.g., ``cup moved from kitchen counter to dining table''). Captions are embedded using Qwen3-Embedding-8B and stored in Text view with links to source object records and temporal spans. Text view enables semantic queries matching object state descriptions via embedding similarity without requiring explicit event boundaries.

\textbf{(ii) Room-level semantic labels:} Scene view octree nodes store semantic room labels inferred by the VLM. These labels are replicated in Text view with spatial references (octree node IDs) and temporal validity ranges, supporting spatial queries phrased in natural language by mapping room names to geometric regions.

\textbf{(iii) Speech transcripts:} Audio streams are transcribed via Whisper-large-v3 and segmented into utterances with speaker labels and timestamps. Raw transcripts are stored in Text view as temporal documents $\{(s_k, [t_k^{\text{start}}, t_k^{\text{end}}], \text{speaker}_k)\}$. This supports queries referencing spoken content through embedding-based retrieval.

Text view maintains a vector index mapping embeddings to source records, enabling efficient nearest-neighbor search across all three content types simultaneously (implemented with FAISS).

\paragraph{Event materialization.}
Events are derived from two sources and stored in the Event view with bidirectional links to participating objects:

\textbf{(i) Object-centric motion events:} For each object $o_i$ with LTE representation, motion phase transitions trigger event creation. When the VLM generates a motion caption $c_j$ for interval $[t_j^{\text{start}}, t_j^{\text{end}}]$, we create an event record $e_j = (d_{\text{description}}, [t_{\text{start}}, t_{\text{end}}], \mathcal{O}_{\text{participants}})$. Participants $\mathcal{O}_{\text{participants}}$ include the primary object and any secondary objects mentioned in the caption (containers, surfaces, manipulated items), resolved via entity linking against the Object view.

\textbf{(ii) Speech-derived activity events:} Audio transcripts are segmented into temporal windows and processed by an LLM (Qwen3-8B) to extract activity-level events. The LLM identifies activities with descriptive text (e.g., ``cooking pasta with tomato sauce''), temporal markers, participants, and mentioned objects. Each extracted event is materialized with descriptive activity text embedded for similarity-based retrieval.

\subsection{LTE Representation Details}
\label{app:lte-details}

\paragraph{Adaptive waypoint selection.}
For motion intervals $[t_j^{\text{start}}, t_j^{\text{end}}]$ with 3D trajectory $\{\mathbf{p}_t\}$, we apply Douglas--Peucker simplification with tolerance $\epsilon = 0.15$\,m. The algorithm recursively identifies the point with maximum perpendicular distance to the line segment connecting endpoints; if this distance exceeds $\epsilon$, the point becomes a waypoint and the trajectory is subdivided. This produces waypoints $\mathcal{A}_{\text{spatial}} = \{\mathbf{p}_{k_1}, \ldots, \mathbf{p}_{k_m}\}$ where $m \ll (t_j^{\text{end}} - t_j^{\text{start}})$ for smooth trajectories.

\paragraph{Visual anchor sampling.}
Visual anchors $\mathcal{A}_{\text{visual}}$ are sampled adaptively based on motion state and trajectory geometry:
\begin{itemize}[nosep,leftmargin=*]
    \item \textbf{Tracking-gap intervals}: store visual anchors only at observable boundaries (last detected frame before the gap, first re-detected frame after the gap if available).
    \item \textbf{Motion intervals}: sample frames at each spatial waypoint from Douglas--Peucker, plus intermediate frames if temporal gap exceeds $10$\,seconds between consecutive waypoints.
    \item \textbf{State transition boundaries}: always sample frames at motion phase boundaries (static $\leftrightarrow$ motion).
\end{itemize}
For each sampled frame where the object is detected, we store the RGB crop within the object's bounding box $b_{t_k}$.

\paragraph{Caption generation.}
For each motion interval, we construct a prompt providing the VLM with: (i) temporally ordered frames sampled at $3$\,FPS from the interval, (ii) object bounding boxes overlaid on frames, (iii) object category name, and (iv) room-level semantic label from Scene view. The VLM generates a concise natural-language description capturing motion pattern and spatial context (see prompt template in Appendix~\ref{app:prompts}).

\subsection{Query Execution Details}
\label{app:query-processing}

Given query $q$, the system executes:

\begin{enumerate}[nosep,leftmargin=*]
   \item \textbf{Parse}: extract structured constraints (objects $\mathcal{O}$, time $\mathcal{T}$, space $\mathcal{R}$, semantics $\mathcal{S}$) using Qwen3-8B. Objects are mapped to entries in the maintained object list via fuzzy string matching, spatial regions are mapped to predefined room categories, and temporal expressions are converted to timestamp ranges or semantic periods (``morning'' $\rightarrow [06{:}00, 12{:}00]$).

   \item \textbf{Object view lookup}: retrieve candidate object records matching $\mathcal{O}$ via (i) exact category matching against object metadata; and (ii) a fallback semantic retrieval using similarity between the query's object description and LTE motion captions (Qwen3-Embedding-8B, threshold $0.75$).

   \item \textbf{Scene view filtering}: apply spatial constraints $\mathcal{R}$ through octree traversal as described in \Cref{sec:memory-representation}.

   \item \textbf{Event view indexing}: for queries involving activities or interactions (detected via parsing), retrieve candidate temporal intervals from Event view where relevant activities occurred. Cross-reference event participants with Object view.

   \item \textbf{Text view matching}: compute semantic similarity between $\mathcal{S}$ and LTE motion captions, retaining captions with $\text{sim}_j > 0.8$.

   \item \textbf{Image view verification}: for visual queries (VQ2D), extract DINOv2 embedding $\mathbf{q} \in \mathbb{R}^{1024}$ from the query image and retain only objects with $\max_k \text{score}_k > 0.7$.

   \item \textbf{Temporal filtering}: apply $\mathcal{T}$ to LTE interval timestamps; retain intervals overlapping with the query range.

   \item \textbf{Evidence assembly}: for matched objects, collect (i)~trajectory segments with per-frame bounding boxes interpolated from spatial anchors, (ii)~motion captions, (iii)~visual anchors for identity verification, and (iv)~event contexts. Return ranked results based on aggregate similarity scores.
\end{enumerate}

\subsection{Baseline Implementation Details}
\label{app:baselines}

\paragraph{VLM baseline (Q3VL-8B/235B + Grounding-DINO).}
For Qwen3-VL baselines on SMB, we utilize the video clips provided in the EgoLife dataset. Query processing follows four steps: (1)~\emph{Temporal windowing}: identify all clips whose timestamps fall within $\mathcal{T}$. (2)~\emph{VLM examination}: sample frames to construct $3$-second video inputs at $3$\,FPS ($9$ frames per window). The VLM outputs an object description if a match is found, or ``not present'' otherwise. We slide non-overlapping windows across each clip. (3)~\emph{Object localization}: for clips where the VLM reports a match, apply Grounding-DINO with the VLM-generated description as text prompt (box confidence threshold $0.3$, text threshold $0.25$). (4)~\emph{Success determination}: identify the last detected occurrence within the query window and check IoU $\ge 0.3$ with the ground-truth box.

\paragraph{Keyframe Memory (KFMem).}
Following the 3D-Mem paradigm, we (i)~select keyframes via DINOv2 diversity-based clustering ($\sim\!600$ keyframes per $24$\,h), (ii)~back-project detected objects to 3D positions using the same ViPE reconstruction as our system, (iii)~store keyframes with their associated object 3D positions and VLM-generated visual descriptions, and (iv)~at query time, parse the query to extract object/spatial/temporal constraints, retrieve candidate keyframes by description similarity (Qwen3-Embedding-8B) and spatial bounding-box intersection, and return the most recent keyframe satisfying all constraints. This baseline does not perform per-object trajectory tracking; it relies on snapshot retrieval.

\paragraph{VideoAgent.}
We use the public VideoAgent implementation, providing it with the same SMB queries. VideoAgent maintains an event-based memory across processed clips and uses an LLM agent to query its memory; we adopt its default configuration with Qwen3-VL-8B as the underlying VLM for fair comparison.

%%%%%%%%%%%%%% Prompt Templates %%%%%%%%%%%%%%
\section{Prompt Templates}
\label{app:prompts}

We provide the prompt templates used for key LLM/VLM processing stages.

\paragraph{Query parsing (Qwen3).}
\begin{small}
\begin{verbatim}
You are a query parser for a spatial memory system. Given a natural language query,
extract and map constraints to the provided vocabularies.

Object list: {object_list}
Room list: [Bedroom, Living room, Kitchen, Dining room, Bathroom, Other]

Instructions:
1. Objects: Extract object mentions and map each to the closest match in object_list.
   If no match exists, return the original mention.
2. Spatial: Extract spatial references and map to room_list. Use "Other" for outdoor /
   transitional spaces.
3. Temporal: Convert time expressions to either:
   - Semantic periods: "morning" (6:00-12:00), "afternoon" (12:00-18:00),
     "evening" (18:00-22:00), "night" (22:00-6:00)
   - Specific timestamps: [start_time, end_time] in HH:MM format
   - Relative references: "yesterday", "2 hours ago"
4. Semantic: Extract state descriptions about objects or events (e.g., "moved from X to Y",
   "was placed", "being used", "after cooking").

Query: {query}

Output in JSON format:
{
  "objects": [{"mention": ..., "mapped": ...}, ...],
  "spatial": {"raw": ..., "mapped_room": ...},
  "temporal": {"raw": ..., "type": "semantic|timestamp|relative", "value": ...},
  "semantic": {"object_state": ..., "event_context": ...}
}
\end{verbatim}
\end{small}

\paragraph{VLM baseline query processing (Qwen3-VL).}
\begin{small}
\begin{verbatim}
This video clip spans [START_TIME] to [END_TIME] on [DAY].

Query: [QUERY_TEXT]

Task: Determine if the target object described in the query appears in this clip and
matches the specified conditions (state, location, temporal context, etc.).

Instructions:
1. Examine all frames in the provided sequence
2. If the target object is present AND matches all query conditions:
   - Describe the object's appearance, location, and relevant actions/states
   - Specify approximate timestamp(s) within the clip where it appears
3. If the target object is absent OR present but does NOT match the query conditions:
   - Respond with exactly: "not present"

Be precise about state matching: if the query specifies "apple being washed", an apple
sitting on the counter should return "not present".

Response format:
- If match found: "[OBJECT_DESCRIPTION] at [TIMESTAMP] in [LOCATION], [STATE/ACTION]"
- If no match: "not present"
\end{verbatim}
\end{small}

\paragraph{Room-level semantic labeling (Qwen3-VL).}
\begin{small}
\begin{verbatim}
You are analyzing a sequence of {num_frames} frames sampled from a {window_duration}-
second video segment.

Task 1: Classify the room type shown in these frames.
Room categories:
- Bedroom: Contains bed, wardrobe, personal items
- Living room: Contains sofa, TV, open social space
- Kitchen: Contains stove, sink, cooking utensils
- Dining room: Contains dining table, chairs for eating
- Bathroom: Contains toilet, shower, sink for hygiene
- Other: Outdoor spaces, hallways, stairs, or unclear

Previous window classification: {prev_room_label}

Task 2: Identify salient keyframes for memory storage.
Select frames that represent:
- Room entry: First clear view when entering a new space
- Room exit: Last view before transitioning to another space
- Salient in-room events: Notable object interactions, state changes, etc.

Output format (JSON):
{
  "room_label": "<category_name>",
  "keyframe_indices": [<list of frame indices>],
  "transition_detected": <true/false>
}
\end{verbatim}
\end{small}

\paragraph{Object motion caption (Qwen3-VL).}
\begin{small}
\begin{verbatim}
Describe the motion and state changes of the highlighted object [{object_name}] in this
video segment. The object is marked with a bounding box in the frames.

Focus on:
- Movement direction and path (e.g., left to right, from table to shelf)
- Interactions with surfaces or containers
- State transitions (picked up, placed down, opened, closed, etc.)
- Spatial context (which room, near what landmarks)

Provide a concise description in one sentence that captures the essential motion pattern.
\end{verbatim}
\end{small}

\paragraph{Event extraction from speech (Qwen3).}
\begin{small}
\begin{verbatim}
Extract activity events from this transcript segment.

Transcript: {transcript}
Timestamp range: {start_time} - {end_time}
Speaker labels: {speaker_list}

For each distinct event mentioned, identify:
1. Activity type: cooking, cleaning, eating, conversation, entertainment, work, etc.
2. Participants: Names or roles of people involved
3. Objects: Items mentioned or implied
4. Location: Room or area if mentioned
5. Temporal markers: Start/end times or duration if indicated

Output as JSON array:
[{
  "activity": ...,
  "participants": [...],
  "objects": [...],
  "location": ...,
  "time_in_segment": {"start": ..., "end": ...},
  "confidence": "high|medium|low"
}, ...]
Only extract events with at least medium confidence.
\end{verbatim}
\end{small}

%%%%%%%%%%%%%% Limitations and Societal Impact %%%%%%%%%%%%%%
\section{Limitations and Societal Impact}
\label{app:limitations}

\paragraph{Limitations.}
System performance is bounded by upstream component reliability. Object tracking is imperfect in long, cluttered egocentric videos: ID switches, missed detections, and track fragmentation propagate to memory construction and can corrupt both trajectory anchors and event linking. Likewise, point cloud reconstruction via ViPE accumulates drift over multi-hour sessions, especially during rapid camera motion and transitions between rooms or floors, which can degrade spatial localization and octree assignment. LTE captions are generated at motion phase boundaries and may miss fine-grained state changes during continuous manipulation. Our system represents objects as atomic entities and does not explicitly model part-level dynamics (door angles, drawer extension, articulated states); this limitation is shared by all object-centric memory systems and extending LTE to articulated parts is an important future direction. Finally, the current system assumes a single persistent environment; extending to multi-environment scenarios would require robust cross-environment object re-identification and map alignment, which we do not address. All long-horizon evaluation is conducted on EgoLife (home environments, $6$ participants); validation on office, warehouse, and outdoor environments is needed for stronger generalisation claims.

\paragraph{Societal impact.}
Long-horizon spatial memory systems store detailed activity records, raising privacy considerations for deployment in shared spaces. We emphasize that our work focuses on first-person, user-owned scenarios where the tracked individual controls their data. Intended applications include assistive technology for users with memory impairments and personal productivity tools. Deployment in shared environments would require explicit consent mechanisms and user-controlled data retention policies. We advocate for responsible development practices that prioritize user autonomy and transparent data handling.

%% file: example_paper.bib
@article{bai2025qwen3,
  title={Qwen3-vl technical report},
  author={Bai, Shuai and Cai, Yuxuan and Chen, Ruizhe and Chen, Keqin and Chen, Xionghui and Cheng, Zesen and Deng, Lianghao and Ding, Wei and Gao, Chang and Ge, Chunjiang and others},
  journal={arXiv preprint arXiv:2511.21631},
  year={2025}
}

@inproceedings{werby2024hierarchical,
  title={Hierarchical open-vocabulary 3d scene graphs for language-grounded robot navigation},
  author={Werby, Abdelrhman and Huang, Chenguang and B{\"u}chner, Martin and Valada, Abhinav and Burgard, Wolfram},
  booktitle={First Workshop on Vision-Language Models for Navigation and Manipulation at ICRA 2024},
  year={2024}
}

@inproceedings{khosla2025relocate,
  title={Relocate: A simple training-free baseline for visual query localization using region-based representations},
  author={Khosla, Savya and Schwing, Alexander and Hoiem, Derek and others},
  booktitle={Proceedings of the Computer Vision and Pattern Recognition Conference},
  pages={3697--3706},
  year={2025}
}

@inproceedings{fan2025prvql,
  title={Prvql: Progressive knowledge-guided refinement for robust egocentric visual query localization},
  author={Fan, Bing and Feng, Yunhe and Tian, Yapeng and Liang, James Chenhao and Lin, Yuewei and Huang, Yan and Fan, Heng},
  booktitle={Proceedings of the IEEE/CVF International Conference on Computer Vision},
  pages={5156--5165},
  year={2025}
}

@inproceedings{vqloc,
  title={Where is my wallet? modeling object proposal sets for egocentric visual query localization},
  author={Xu, Mengmeng and Li, Yanghao and Fu, Cheng-Yang and Ghanem, Bernard and Xiang, Tao and P{\'e}rez-R{\'u}a, Juan-Manuel},
  booktitle={Proceedings of the IEEE/CVF Conference on Computer Vision and Pattern Recognition},
  pages={2593--2603},
  year={2023}
}

@article{feng2025osgnet,
  title={OSGNet@ Ego4D Episodic Memory Challenge 2025},
  author={Feng, Yisen and Zhang, Haoyu and Chu, Qiaohui and Liu, Meng and Guan, Weili and Wang, Yaowei and Nie, Liqiang},
  journal={arXiv preprint arXiv:2506.03710},
  year={2025}
}

@article{hou2023groundnlq,
  title={Groundnlq@ ego4d natural language queries challenge 2023},
  author={Hou, Zhijian and Ji, Lei and Gao, Difei and Zhong, Wanjun and Yan, Kun and Li, Chao and Chan, Wing-Kwong and Ngo, Chong-Wah and Duan, Nan and Shou, Mike Zheng},
  journal={arXiv preprint arXiv:2306.15255},
  year={2023}
}

@article{pei2024egovideo,
  title={Egovideo: Exploring egocentric foundation model and downstream adaptation},
  author={Pei, Baoqi and Chen, Guo and Xu, Jilan and He, Yuping and Liu, Yicheng and Pan, Kanghua and Huang, Yifei and Wang, Yali and Lu, Tong and Wang, Limin and others},
  journal={arXiv preprint arXiv:2406.18070},
  year={2024}
}

@inproceedings{whisper,
  title={Robust speech recognition via large-scale weak supervision},
  author={Radford, Alec and Kim, Jong Wook and Xu, Tao and Brockman, Greg and McLeavey, Christine and Sutskever, Ilya},
  booktitle={International conference on machine learning},
  pages={28492--28518},
  year={2023},
  organization={PMLR}
}

@inproceedings{groundingdino,
  title={Grounding dino: Marrying dino with grounded pre-training for open-set object detection},
  author={Liu, Shilong and Zeng, Zhaoyang and Ren, Tianhe and Li, Feng and Zhang, Hao and Yang, Jie and Jiang, Qing and Li, Chunyuan and Yang, Jianwei and Su, Hang and others},
  booktitle={European conference on computer vision},
  pages={38--55},
  year={2024},
  organization={Springer}
}

@article{qwen3embedding,
  title={Qwen3 Embedding: Advancing Text Embedding and Reranking Through Foundation Models},
  author={Zhang, Yanzhao and Li, Mingxin and Long, Dingkun and Zhang, Xin and Lin, Huan and Yang, Baosong and Xie, Pengjun and Yang, An and Liu, Dayiheng and Lin, Junyang and others},
  journal={arXiv preprint arXiv:2506.05176},
  year={2025}
}

@article{tschannen2025siglip2,
  title={Siglip 2: Multilingual vision-language encoders with improved semantic understanding, localization, and dense features},
  author={Tschannen, Michael and Gritsenko, Alexey and Wang, Xiao and Naeem, Muhammad Ferjad and Alabdulmohsin, Ibrahim and Parthasarathy, Nikhil and Evans, Talfan and Beyer, Lucas and Xia, Ye and Mustafa, Basil and others},
  journal={arXiv preprint arXiv:2502.14786},
  year={2025}
}

@article{oquab2023dinov2,
  title={Dinov2: Learning robust visual features without supervision},
  author={Oquab, Maxime and Darcet, Timoth{\'e}e and Moutakanni, Th{\'e}o and Vo, Huy and Szafraniec, Marc and Khalidov, Vasil and Fernandez, Pierre and Haziza, Daniel and Massa, Francisco and El-Nouby, Alaaeldin and others},
  journal={arXiv preprint arXiv:2304.07193},
  year={2023}
}

@article{yang2025qwen3,
  title={Qwen3 technical report},
  author={Yang, An and Li, Anfeng and Yang, Baosong and Zhang, Beichen and Hui, Binyuan and Zheng, Bo and Yu, Bowen and Gao, Chang and Huang, Chengen and Lv, Chenxu and others},
  journal={arXiv preprint arXiv:2505.09388},
  year={2025}
}

@article{sam3,
  title={Sam 3: Segment anything with concepts},
  author={Carion, Nicolas and Gustafson, Laura and Hu, Yuan-Ting and Debnath, Shoubhik and Hu, Ronghang and Suris, Didac and Ryali, Chaitanya and Alwala, Kalyan Vasudev and Khedr, Haitham and Huang, Andrew and others},
  journal={arXiv preprint arXiv:2511.16719},
  year={2025}
}

@article{yang2023set,
  title={Set-of-mark prompting unleashes extraordinary visual grounding in gpt-4v},
  author={Yang, Jianwei and Zhang, Hao and Li, Feng and Zou, Xueyan and Li, Chunyuan and Gao, Jianfeng},
  journal={arXiv preprint arXiv:2310.11441},
  year={2023}
}

@article{huang2025vipe,
  title={Vipe: Video pose engine for 3d geometric perception},
  author={Huang, Jiahui and Zhou, Qunjie and Rabeti, Hesam and Korovko, Aleksandr and Ling, Huan and Ren, Xuanchi and Shen, Tianchang and Gao, Jun and Slepichev, Dmitry and Lin, Chen-Hsuan and others},
  journal={arXiv preprint arXiv:2508.10934},
  year={2025}
}

@inproceedings{yang2025egolife,
  title={Egolife: Towards egocentric life assistant},
  author={Yang, Jingkang and Liu, Shuai and Guo, Hongming and Dong, Yuhao and Zhang, Xiamengwei and Zhang, Sicheng and Wang, Pengyun and Zhou, Zitang and Xie, Binzhu and Wang, Ziyue and others},
  booktitle={Proceedings of the Computer Vision and Pattern Recognition Conference},
  pages={28885--28900},
  year={2025}
}

@article{yan2013semantic,
  title={Semantic trajectories: Mobility data computation and annotation},
  author={Yan, Zhixian and Chakraborty, Dipanjan and Parent, Christine and Spaccapietra, Stefano and Aberer, Karl},
  journal={ACM Transactions on Intelligent Systems and Technology (TIST)},
  volume={4},
  number={3},
  pages={1--38},
  year={2013},
  publisher={ACM New York, NY, USA}
}

@INPROCEEDINGS{schmid2022panoptic,
  author={Schmid, Lukas and Delmerico, Jeffrey and Schönberger, Johannes L. and Nieto, Juan and Pollefeys, Marc and Siegwart, Roland and Cadena, Cesar},
  booktitle={2022 International Conference on Robotics and Automation (ICRA)}, 
  title={Panoptic Multi-TSDFs: a Flexible Representation for Online Multi-resolution Volumetric Mapping and Long-term Dynamic Scene Consistency}, 
  year={2022},
  volume={},
  number={},
  pages={8018-8024},
  doi={10.1109/ICRA46639.2022.9811877}
}

@inproceedings{sarch2023helper,
  title={Open-ended instructable embodied agents with memory-augmented large language models},
  author={Sarch, Gabriel and Wu, Yue and Tarr, Michael and Fragkiadaki, Katerina},
  booktitle={Findings of the Association for Computational Linguistics: EMNLP 2023},
  pages={3468--3500},
  year={2023}
}

@article{rosinol2021kimera,
  title={Kimera: From SLAM to spatial perception with 3D dynamic scene graphs},
  author={Rosinol, Antoni and Violette, Andrew and Abate, Marcus and Hughes, Nathan and Chang, Yun and Shi, Jingnan and Gupta, Arjun and Carlone, Luca},
  journal={The International Journal of Robotics Research},
  volume={40},
  number={12-14},
  pages={1510--1546},
  year={2021},
  publisher={SAGE Publications Sage UK: London, England}
}

@inproceedings{videoagent2024,
  title={Videoagent: A memory-augmented multimodal agent for video understanding},
  author={Fan, Yue and Ma, Xiaojian and Wu, Rujie and Du, Yuntao and Li, Jiaqi and Gao, Zhi and Li, Qing},
  booktitle={European Conference on Computer Vision},
  pages={75--92},
  year={2024},
  organization={Springer}
}

@inproceedings{amego2024,
  title={Amego: Active memory from long egocentric videos},
  author={Goletto, Gabriele and Nagarajan, Tushar and Averta, Giuseppe and Damen, Dima},
  booktitle={European Conference on Computer Vision},
  pages={92--110},
  year={2024},
  organization={Springer}
}

@inproceedings{pramanick2023egovlpv2,
  title={Egovlpv2: Egocentric video-language pre-training with fusion in the backbone},
  author={Pramanick, Shraman and Song, Yale and Nag, Sayan and Lin, Kevin Qinghong and Shah, Hardik and Shou, Mike Zheng and Chellappa, Rama and Zhang, Pengchuan},
  booktitle={Proceedings of the IEEE/CVF International Conference on Computer Vision},
  pages={5285--5297},
  year={2023}
}

@article{lin2022egovlp,
  title={Egocentric video-language pretraining},
  author={Lin, Kevin Qinghong and Wang, Jinpeng and Soldan, Mattia and Wray, Michael and Yan, Rui and Xu, Eric Z and Gao, Difei and Tu, Rong-Cheng and Zhao, Wenzhe and Kong, Weijie and others},
  journal={Advances in Neural Information Processing Systems},
  volume={35},
  pages={7575--7586},
  year={2022}
}

@inproceedings{grauman2022ego4d,
  title={Ego4d: Around the world in 3,000 hours of egocentric video},
  author={Grauman, Kristen and Westbury, Andrew and Byrne, Eugene and Chavis, Zachary and Furnari, Antonino and Girdhar, Rohit and Hamburger, Jackson and Jiang, Hao and Liu, Miao and Liu, Xingyu and others},
  booktitle={Proceedings of the IEEE/CVF conference on computer vision and pattern recognition},
  pages={18995--19012},
  year={2022}
}

@article{zheng2015trajectory,
  title={Trajectory data mining: an overview},
  author={Zheng, Yu},
  journal={ACM Transactions on Intelligent Systems and Technology (TIST)},
  volume={6},
  number={3},
  pages={1--41},
  year={2015},
  publisher={ACM New York, NY, USA}
}

@article{douglas1973algorithms,
  title={Algorithms for the reduction of the number of points required to represent a digitized line or its caricature},
  author={Douglas, David H and Peucker, Thomas K},
  journal={Cartographica: the international journal for geographic information and geovisualization},
  volume={10},
  number={2},
  pages={112--122},
  year={1973},
  publisher={University of Toronto Press}
}

@article{khronos2024,
  title={Khronos: A unified approach for spatio-temporal metric-semantic slam in dynamic environments},
  author={Schmid, Lukas and Abate, Marcus and Chang, Yun and Carlone, Luca},
  journal={arXiv preprint arXiv:2402.13817},
  year={2024}
}

@inproceedings{karma2025,
  title={Karma: Augmenting embodied ai agents with long-and-short term memory systems},
  author={Wang, Zixuan and Yu, Bo and Zhao, Junzhe and Sun, Wenhao and Hou, Sai and Liang, Shuai and Hu, Xing and Han, Yinhe and Gan, Yiming},
  booktitle={2025 IEEE International Conference on Robotics and Automation (ICRA)},
  pages={1--8},
  year={2025},
  organization={IEEE}
}

@inproceedings{gu2024conceptgraphs,
  title={Conceptgraphs: Open-vocabulary 3d scene graphs for perception and planning},
  author={Gu, Qiao and Kuwajerwala, Ali and Morin, Sacha and Jatavallabhula, Krishna Murthy and Sen, Bipasha and Agarwal, Aditya and Rivera, Corban and Paul, William and Ellis, Kirsty and Chellappa, Rama and others},
  booktitle={2024 IEEE International Conference on Robotics and Automation (ICRA)},
  pages={5021--5028},
  year={2024},
  organization={IEEE}
}

@inproceedings{3dmem2025,
  title={3D-mem: 3D scene memory for embodied exploration and reasoning},
  author={Yang, Yuncong and Yang, Han and Zhou, Jiachen and Chen, Peihao and Zhang, Hongxin and Du, Yilun and Gan, Chuang},
  booktitle={Proceedings of the Computer Vision and Pattern Recognition Conference},
  pages={17294--17303},
  year={2025}
}
